%% file: main.tex
\PassOptionsToPackage{safe}{tipa}
\documentclass[cameraready]{Interspeech}

\title{Physics-Aware Deepfake Detection via Distance–Speech Consistency}

\author[affiliation={1}, orcid=0009-0006-6102-5860]{Kyeongrae}{Kim}
\author[affiliation={2}, orcid=0000-0003-3542-9934]{Kim}{Sung-Bin}
\author[affiliation={2}, orcid=0009-0002-1511-7482]{Oh}{Hyun-Bin}
\author[affiliation={3}, orcid=0000-0003-0468-1571]{Tae-Hyun}{Oh}

\address{
    $^1$School of Electrical Engineering and $^3$School of Computing, KAIST, Korea \\
    $^{2}$Dept. of Electrical Engineering, POSTECH, Korea 
}

\email{kimkyeongrae@kaist.ac.kr, \{sungbin, hyunbinoh\}@postech.ac.kr, taehyun.oh@kaist.ac.kr}


\usepackage{booktabs}   
\usepackage{multirow}   
\usepackage{graphicx}   
\usepackage{tabularx}
\usepackage{float} 
\usepackage{comment}
\input{def}

\begin{document}

\maketitle

\begin{abstract}
    Recent progress in generative models has made talking-head deepfakes highly realistic and easy to create, jeopardizing public trust. However, most audio-visual deepfake detectors rely on lip–speech synchronization and are designed for static, frontal videos, limiting their reliability in dynamic, in-the-wild recordings where the speaker moves and lip cues are degraded or missing. As an alternative, we propose a physics-aware detector for dynamic speaking videos that leverages an acoustic constraint: in real recordings, speech energy varies predictably with speaker-to-camera distance, whereas deepfake generation can break this coupling. Our method combines distance estimates from video with distance-dependent acoustic measures from speech to detect deepfakes. Experiments on in-the-wild benchmarks show that the proposed physical cue is effective in dynamic scenarios, and that it complements lip-sync-based detectors, with a simple ensemble achieving consistent gains across datasets. Codes are available at \url{https://byulharang.github.io/PADD/}.

\end{abstract}

\section{Introduction}
\input{1_Intro}

\section{Method}
\input{2_Method}

\section{Experiments}
\input{3_Experiments}
\section{Conclusion}
\input{4_Conclusion}

\section{Acknowledgments}
\input{5_Acknowledgements}

\section{Use of Generative AI Disclosure}
\input{6_Generative_AI_Disclosure}

\bibliographystyle{IEEEtran}
\bibliography{mybib}

\end{document}

%% file: def.tex
\usepackage{enumitem}
\usepackage{multirow}
\usepackage{wrapfig}
\usepackage{colortbl}
\usepackage[normalem]{ulem}
\usepackage{microtype}
\usepackage{comment}
\usepackage[hang,flushmargin]{footmisc}
\usepackage{graphicx}
\usepackage{multirow}
\usepackage{pifont}
\usepackage{nicematrix}
\usepackage{ulem}
\usepackage{diagbox} 
\usepackage{tipa}

\setlist[itemize]{align=parleft,left=0pt,topsep=1mm,itemsep=0mm,parsep=1mm}
\definecolor{azure}{rgb}{0.0, 0.5, 1.0}
\definecolor{azure(colorwheel)}{rgb}{0.0, 0.5, 1.0}
\definecolor{nicegreen}{rgb}{0.0, 0.7, 0.1}
\definecolor{yw}{rgb}{0.01176, 0.5490, 0.5490}
\definecolor{ashblue}{rgb}{0.36, 0.54, 0.66}
\definecolor{ashgrey}{rgb}{0.7, 0.75, 0.71}
\definecolor{applegreen}{rgb}{0.55, 0.71, 0.0}
\definecolor{blue}{rgb}{0.0, 0.0, 1.0}
\definecolor{red}{rgb}{0.6, 0.0, 0.12}
\definecolor{postechred}{rgb}{0.784, 0.003, 0.313}
\definecolor{ywg}{rgb}{0.9960, 0.8984, 0.5859}
\definecolor{ballblue}{rgb}{0.13, 0.67, 0.8}
\definecolor{cornellred}{rgb}{0.7, 0.11, 0.11}
\definecolor{darkcyan}{rgb}{0.0, 0.55, 0.55}
\definecolor{CuGray}{gray}{0.9}
\definecolor{airforceblue}{rgb}{0.36, 0.54, 0.66}
\definecolor{rev}{rgb}{0.784, 0.003, 0.313}
\definecolor{pink}{cmyk}{0, 0.7808, 0.4429, 0.1412}
\definecolor{amethyst}{rgb}{0.6, 0.4, 0.8}
\definecolor{black}{rgb}{0.0, 0.0, 0.0}
\definecolor{tb3_yellow}{rgb}{0.996, 1.0, 0.6}
\definecolor{tb3_orange}{rgb}{0.980, 0.8, 0.604}
\definecolor{tb3_red}{rgb}{0.972, 0.6, 0.6}
\definecolor{dimgray}{rgb}{0.41, 0.41, 0.41}
\definecolor{brickred}{rgb}{0.8, 0.25, 0.33}
\definecolor{bleudefrance}{rgb}{0.19, 0.55, 0.91}
\definecolor{blue(ncs)}{rgb}{0.265, 0.445, 0.765}
\definecolor{blue(ryb)}{rgb}{0.01, 0.28, 1.0}
\definecolor{orange}{rgb}{1.0, 0.49, 0.0}
\definecolor{Gray}{gray}{0.88}
\definecolor{green(ncs)}{rgb}{0.0, 0.62, 0.42}
\definecolor{clova}{rgb}{0.24, 0.63, 0.33}
\definecolor{correctgreen}{rgb}{0.01, 0.702, 0.321}
\definecolor{wrongred}{rgb}{0.757, 0, 0}

\definecolor{teaserblue}{rgb}{0.274, 0.694, 1}
\definecolor{teaserorange}{rgb}{0.913, 0.443, 0.196}
\definecolor{teaserred}{rgb}{1,0,0}

\definecolor{kellygreen}{rgb}{0.3, 0.73, 0.09}

\newcolumntype{g}{>{\columncolor{CuGray}}c}
\newcolumntype{z}{>{\columncolor{CuGray}}l}

\renewcommand{\paragraph}[1]{\noindent\textbf{#1{.}}\,}

\definecolor{steelblue}{rgb}{0.27, 0.51, 0.7}

\usepackage{xspace}

\makeatletter
\def\@fnsymbol#1{\ensuremath{\ifcase#1\or *\or \dagger\or \ddagger\or
   \mathsection\or \mathparagraph\or \|\or **\or \dagger\dagger
   \or \ddagger\ddagger \else\@ctrerr\fi}}
\makeatother

\def\onedot{.\@\xspace}
\def\eg{\emph{e.g}\onedot} 
\def\ie{\emph{i.e}\onedot}

\newcommand{\be}{\begin{eqnarray}}
\newcommand{\ee}{\end{eqnarray}}
\newcommand{\bee}{\begin{eqnarray*}}
\newcommand{\eee}{\end{eqnarray*}}

\newcommand{\matrixb}{\left[ \begin{array}}
\newcommand{\matrixe}{\end{array} \right]}

\definecolor{graygray}{HTML}{bbbbbb}
 
\usepackage{mathtools}

\definecolor{ExtFG}{HTML}{7B2CBF}   
\definecolor{OrigFG}{HTML}{A44A3F}  

\usepackage{booktabs}     
\usepackage{nicematrix}   
\usepackage{xcolor} 
\usepackage{graphicx} 
\definecolor{defaultColor}{RGB}{230, 244, 252}
\definecolor{baselineColor}{RGB}{245,245,245}
\definecolor{oursColor}{RGB}{230,244,252} 
\definecolor{ensColor}{RGB}{236,248,240}
\usepackage{xcolor}
\usepackage{subcaption}
\definecolor{distcolor}{RGB}{55,126,184}   
\definecolor{snrcolor}{RGB}{77,175,74}     
\definecolor{c50color}{RGB}{228,26,28}     

\expandafter\def\expandafter\normalsize\expandafter{%
    \normalsize%
    \setlength{\abovedisplayskip}{6pt}
    \setlength{\belowdisplayskip}{6pt}
    \setlength{\abovedisplayshortskip}{6pt}
    \setlength{\belowdisplayshortskip}{6pt}
}
\usepackage{placeins}
\usepackage{dblfloatfix}

%% file: 1_Intro.tex
Recent advances in deep generative models~\cite{polyak2024movie, google2024veo, wiedemer2025video} have dramatically increased the realism and accessibility of synthetic media, raising the risk of convincing deepfake videos in everyday communication channels and public discourse~\cite{gong2024contemporary, xie2025deepfake}. 
To defense these threats, various detection methods have been developed for video forensics, ranging from detection of visual artifacts~\cite{shiohara2022detecting, yang2025d, rossler2019faceforensics++} to verification of temporal and spatial consistency~\cite{guo2025face, nguyen2025vulnerability, zheng2025d3}.
In particular, manipulating a human speech video is especially threatening 
since it can distort identity (visual appearance) and intent (speech content), and may influence perceived beliefs and decisions even with a short clip, exacerbating misinformation and jeopardizing public trust.
This highlights the need for robust audio-visual deepfake detection.

\begin{figure}[t] 
  \centering
  \includegraphics[width=\linewidth]{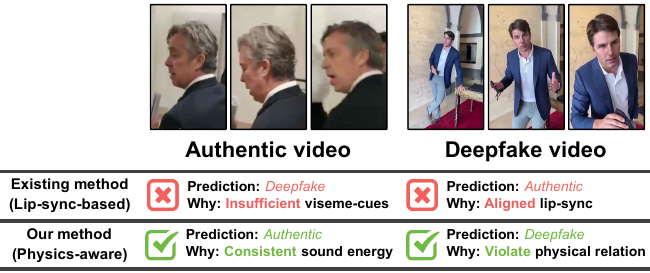}
    \vspace{-5mm}
    \caption{
    \textbf{Physics-aware deepfake detection for dynamic speaking videos.}
    Existing detectors rely on lip–speech consistency and may fail when lip regions are partially visible or missing.
    Our method instead exploits the physical relationship between speaker distance and speech energy, enabling reliable detection even when visual lip cues are unreliable.
    }
    \vspace{-6mm}
  \label{fig:teaser}
\end{figure}

\begin{figure*}[!t] 
  \centering
  \includegraphics[width=0.88\linewidth]{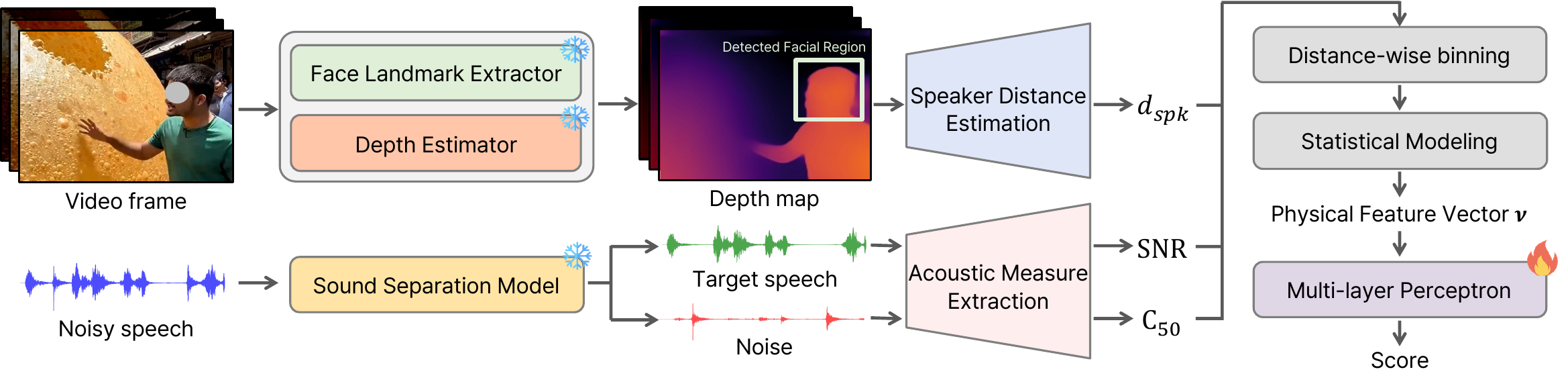}
    \vspace{-4mm}
    \caption{
    \textbf{Physics-aware deepfake detection framework.}
    Our method extracts speaker distance and acoustic measures from video and audio streams.
    For speaker distance, we calculated $d_{spk}$ by averaging monocular depth values (MiDaS~\cite{ranftl2020towards}) within facial landmarks (RetinaFace~\cite{deng2020retinaface}). For acoustic measures, SNR and $C_{50}$ were calculated using speech and noise separated by SAM-Audio~\cite{shi2025sam} with a 'speech' prompt.
    Distance-wise binning is then applied and the variance and entropy of the acoustic measures are aggregated to form statistical features $\mathbf{v}$.
    Finally, $\mathbf{v}$ is fed into a lightweight MLP detector to predict whether it is an authentic or a deepfake video.
    }
  \vspace{-5mm}
  \label{fig:method}
\end{figure*}

Most existing audio-visual deepfake detection methods have primarily exploited the strong correlation between speech and lip motion, analyzing lip-speech synchronization to detect subtle mismatches~\cite{astrid2024detecting, zou2024cross, Zhou_2021_ICCV,liang2024speechforensics, koutlis2026auvire}.
These approaches have shown to be effective in videos where the speaker's mouth is clearly visible and unobstructed, such as news, interviews, podcasts, and studio-like recordings~\cite{khalid2021fakeavceleb,cai2024av,rossler2019faceforensics++,cai2023glitch}.
However, this reliance on lip–speech synchronization limits practical deployment to relatively constrained settings, such as static talking head videos with frontal faces, stable lighting, and limited camera motion, where the lip region remains consistently visible.

As generative tools rapidly advance while becoming cheaper and easier to use, deepfakes are no longer limited to carefully curated static talking head clips; instead, they increasingly appear in dynamic, in-the-wild videos where a subject moves through space while speaking, such as handheld recordings, vlogs, or on-the-street interviews.
While effective under controlled and static conditions, the aforementioned approaches become unreliable in dynamic speaking videos, where the speaker naturally moves toward or away from the camera and facial cues are often degraded by occlusion, pose changes, and motion blur.

To address this gap, we propose a physics-aware deepfake detection framework for dynamic human speaking videos. Our key motivation is that authentic recordings must satisfy basic acoustic constraints: speech intensity varies predictably with the speaker-to-camera distance, approximately following an inverse relationship~\cite{kinsler2000fundamentals}. 
Although current deepfake generation and compositing methods can produce highly realistic content, they often neglect this physical relationship and may break the coupling between visual geometry and speech energy. 
Our approach estimates the speaker's relative distance from the visual stream and compares it against speech-energy patterns expected from sound propagation. This provides a complementary detection signal that remains informative even when lip details are unreliable, and targets violations of the physical relationship between what is seen (distance/motion) and what is heard (speech intensity) in dynamic deepfake videos (refer to Fig.~\ref{fig:teaser}).

We validate the effectiveness of our approach on multiple audio-visual deepfake detection benchmarks that include dynamic videos~\cite{koutlis2026auvire, chandra2025deepfake}. When we explicitly extract dynamic subsets from these benchmarks and evaluate on those videos, our physics-aware model outperforms prior detectors that primarily focus on lip–speech alignment~\cite{liang2024speechforensics, koutlis2026auvire}. Moreover, our method performs favorably as a standalone detector compared to existing approaches. Because our physical-consistency cue complements prior detectors, ensembling it with existing models achieves substantial additional gains.
To our knowledge, this is the first work to use audio-visual physical constraints as a primary cue for deepfake video detection. Our main contributions are:
\begin{itemize}
    \item Introduce a distance–speech consistency measure that reflects basic acoustic behavior in real dynamic speaking videos.
    \item Propose an audio-visual deepfake detector that leverages this measure as a primary cue for detection in dynamic settings.
    \item Demonstrate robust performance on dynamic speaking videos and consistent ensemble gains with lip–speech synchronization detectors, indicating complementarity.
\end{itemize}

%% file: 2_Method.tex


This section presents our physics-aware deepfake detection framework based on \emph{physical alignment} between the speaker's visual distance and speech acoustics.
We first introduce the acoustic principle and distance-related measures used in our method (Sec.~\ref{sec:preliminaries}).
We then describe how we extract acoustic measures and relative speaker distance, and how we convert them into robust video-level features by distance-wise binning and distributional statistics, \eg, variance and entropy (Sec.~\ref{sec:extract}).
Finally, we build a lightweight multi-layer perceptron (MLP) that maps these statistics to a deepfake score (Sec.~\ref{sec:detector}).

\subsection{Preliminaries}
\label{sec:preliminaries}

In real-world recordings, a speaker's distance and speech signals are coupled through acoustic propagation: as distance increases, speech intensity decreases, and vice versa~\cite{levkovitch2024zero, patterson2022distance}. This is known as the inverse-square law. Based on this principle, we use two acoustic measures, signal-to-noise ratio (SNR) and clarity at 50 ms ($C_{50}$), to capture distance-dependent behavior and assess physical consistency between spatial and acoustic signals.

\paragraph{Inverse-square law}
In acoustics, the inverse-square law describes the physical relationship between sound intensity and the source-receiver distance~\cite{kinsler2000fundamentals}. 
Assuming a source radiates power $P$ uniformly, the acoustic intensity $I$ at distance $d$ follows:
$$I = \frac{P}{4\pi d^2} \propto \frac{1}{d^2}.$$
Directly applying the inverse-square law to real-world recordings is often unreliable due to automatic gain control (AGC), device-specific amplification, and recording-level normalization. We therefore employ \textit{relative energy measures}~\cite{kinsler2000fundamentals, oppenheim1999discrete} that are robust to these effects.
Specifically, we utilize the Signal-to-Noise Ratio ($\text{SNR}$) and Clarity at 50ms (${C}_{50}$), which capture ratios of energy rather than absolute magnitudes and thus reflect the inverse-square attenuation of direct sound.
This makes them particularly suitable for in-the-wild videos where calibrated signals or ground-truth acoustic measurements are typically unavailable.

\paragraph{Signal-to-Noise Ratio (SNR)}
SNR measures the relative energy between the target source signal and environmental noise~\cite{oppenheim1999discrete}. 
Let $S_{\text{received}}$ denotes the speech signal captured by the microphone and $N_{\text{env}}$ denotes environmental noise.
In typical recording conditions, environmental noise varies more slowly than the direct speech component within a short temporal window. 
Under this approximation, the variation of SNR is primarily governed by the attenuation of the direct sound with distance.
Thus, when the direct part of speech signal follows the inverse-square law, the SNR of the received signal approximately follows:
$$
\text{SNR~[dB]} \approx 10 \log_{10} \left( \frac{\mathbb{E}[S_{\text{received}}^2]}{\mathbb{E}[N_{\text{env}}^2]} \right) \propto -20 \log_{10}(r).
$$

\paragraph{Clarity at 50ms ($C_{50}$)}
$C_{50}$ measures speech clarity by computing the decibel ratio between the energy arriving within the first 50ms (direct sound and early reflections) and the energy arriving afterward (late reverberation) in the received signal~\cite{kuttruff2016room}. Because the early component is typically dominated by the direct sound, which approximately follows inverse-square attenuation of the source energy ($E_{\text{source}}/r^{2}$), $C_{50}$ can be approximated as:
$$
C_{50}~\text{[dB]} \approx 10 \log_{10} \left( \frac{E_{\text{source}} / r^2}{E_{\text{late}}} \right)
\propto -20 \log_{10}(r),
$$
where $E_{\text{late}}$ denotes the energy of late reflections.
Although $C_{50}$ is commonly used in room acoustics, the same intuition extends to outdoor recordings where the signal still contains a direct propagation component~\cite{kuttruff2016room, kang2000sound}. Consequently, $C_{50}$ provides a physically grounded cue for the source–receiver distance.

\subsection{Extracting Distance-dependent Acoustic Features}
\label{sec:extract}
Our framework extracts physically meaningful features by jointly using acoustic propagation and visual distance cues (see Fig.~\ref{fig:method}).
We derive acoustic measures (SNR and $C_{50}$) from speech signals and estimate the relative speaker distance from video frames.
These signals are then aggregated through distance-wise binning and statistical modeling to obtain video-level features.


\paragraph{Acoustic measure extraction}
We extract SNR and $C_{50}$ from the speech signal. 
To compute these, the speech component must be separated from environmental noise.
Thus, we employ SAM Audio~\cite{shi2025sam} with the text prompt ``speech'' to isolate the speech signal from the input audio.
Using the separated signals, we compute SNR and $C_{50}$ over temporal windows of 300\,ms, which covers typical syllable durations in speech~\cite{poeppel2003analysis} while capturing most of  reflections associated with room reverberation~\cite{diaz2005reverberation}.
This design enables stable estimation of the acoustic measures while reducing interference from adjacent utterances.

\paragraph{Speaker distance estimation}
We estimate the relative distance between the speaker and the camera for each frame. 
Facial landmarks are detected using RetinaFace~\cite{deng2020retinaface} to localize the speaker region. 
A dense depth map of the frame is estimated using MiDaS~\cite{ranftl2020towards}. 
We then obtain speaker distance by averaging the depth values within the detected facial region, and its inverse is used as a proxy for the speaker's relative distance $d_{spk}$.

\paragraph{Distance-wise binning}
We use the speaker distance $d_{spk}$ to group acoustic measures into distance bins.
Specifically, a distance range within each video is partitioned into the bins, and acoustic measures are averaged within each bin.
This reduces fluctuations in vocal intensity and phonetic content that may violate the constant-power assumption of the physical model, allowing the acoustic measures to follow an attenuation trend:
\[
\mathbb{E}[\{\mathrm{SNR}, {C}_{50}\} \mid d_{spk}] \propto -20 \log_{10}(d_{spk}).
\]

\paragraph{Statistical modeling of acoustic measures}
Although the expected acoustic measures follow the physical attenuation trend, instantaneous values of SNR and $C_{50}$ are sensitive to speech variability and environmental noise. 
Thus, we characterize the distribution of acoustic measures within each depth bin using variance and entropy, which capture the following relationships.

\textit{Variance.} When speech energy appears inconsistent with the speaker's visual distance, as often observed in deepfake videos, acoustic measures within the same depth bin become more dispersed, resulting in larger variance. 
In contrast, authentic recordings show lower variance since acoustic propagation remains physically consistent with speaker distance:
$Var[x\!\mid\!d_{spk}]_{\textrm{Fake}} >
Var[x\!\mid\!d_{spk}]_{\textrm{Real}}$, where $x\in\{\mathrm{SNR},{C}_{50}\}$.


\textit{Entropy.} In deepfake videos, synthesized speech often exhibits reduced variability in acoustic measures, resulting in lower entropy~\cite{ren2022revisiting}. In contrast, real-world videos are influenced by environmental noise and unconstrained recording conditions~\cite{bellows2022effect}, leading to higher entropy:
$\mathrm{H}(x\mid d_{spk})_{\textrm{Real}} >
\mathrm{H}(x\mid d_{spk})_{\textrm{Fake}}$.
We aggregate the bin-wise variance and entropy of acoustic measures to obtain video-level statistics:
\[
\begin{aligned}
\mathbb{E}_{\textrm{bin}}[Var(x\mid d_{spk})]_{\textrm{Fake}} &>
\mathbb{E}_{\textrm{bin}}[Var(x\mid d_{spk})]_{\textrm{Real}} \\
\mathbb{E}_{\textrm{bin}}[\mathrm{H}(x\mid d_{spk})]_{\textrm{Real}} &>
\mathbb{E}_{\textrm{bin}}[\mathrm{H}(x\mid d_{spk})]_{\textrm{Fake}}
\end{aligned}
\]
where $\mathbb{E}_{\textrm{bin}}$ denotes the expectation over depth bins.
\vspace{-1mm}

\subsection{Physics-aware Deepfake Detector}
\label{sec:detector}
To capture the statistical behavior of acoustic signals across distance bins, we extract features based on the variance and entropy of SNR and $C_{50}$.
For each acoustic measure $x \in \{\mathrm{SNR}, C_{50}\}$, we compute the expected variance and expected entropy across distance bins, denoted as
$\mathbb{E}_{\textrm{bin}}[Var(x|d_{spk})]$ and $\mathbb{E}_{\textrm{bin}}[\mathrm{H}(x|d_{spk})]$, respectively.
These statistics capture how acoustic properties vary with respect to the estimated speaker distance. Using these statistics, we define the following features:
\[
Var_{\textrm{SNR}} \! = \mathbb{E}_{\textrm{bin}}[Var(\mathrm{SNR}|d_{spk})],
Var_{C_{50}}\!=\mathbb{E}_{\textrm{bin}}[Var(C_{50}|d_{spk})],
\vspace{-3mm}
\]
\[
\mathrm{H}_{\textrm{SNR}}\!=\mathbb{E}_{\textrm{bin}}[\mathrm{H}(\mathrm{SNR}|d_{spk})],
\mathrm{H}_{C_{50}}\!=\mathbb{E}_{\textrm{bin}}[\mathrm{H}(C_{50}|d_{spk})].
\]
To further capture the relative variability between acoustic features, we additionally compute the variance ratio:
\[
r_{\textrm{var}} =
\frac{Var_{\textrm{SNR}}}{Var_{C_{50}}}.
\vspace{-1mm}
\]
We then construct a five-dimensional feature vector:
\[
\mathbf{v} =
\left[
Var_{\textrm{SNR}},
Var_{C_{50}},
\mathrm{H}_{\textrm{SNR}},
\mathrm{H}_{C_{50}},
r_{var}
\right].
\]
Finally, we train a multi-layer perceptron (MLP) $F$ with a single hidden layer that maps $\mathbf{v}$ to a scalar score $F(\mathbf{v}) \in \mathbb{R}$, where scores closer to $1$ indicate fake and closer to $0$ indicate real.
where $\mathbf{x}\in\{SNR, C_{50}\}$ and $\mathbb{E}_{bin}$.

%% file: 3_Experiments.tex
In this section, we compare our physics-aware detector with state-of-the-art lip-sync-based detectors. 
We first evaluate the models on a curated dynamic subset, where physical cues are more informative.
Then, we report ROC--AUC scores of the detectors on full set (\ie, both static and dynamic) of in-the-wild benchmarks. 
We also present qualitative examples and ablation studies to further analyze the method.

\subsection{Experiment Setup}

\paragraph{Competing methods}
We compare against two state-of-the-art lip-sync-based deepfake detectors, SpeechForensics~\cite{liang2024speechforensics} and AuViRe~\cite{koutlis2026auvire}. 
Both methods focus on the alignment between lip movements and speech, and adopt AV-HuBERT~\cite{shi2022learning} as the backbone encoder for audio-visual representation learning.
They have demonstrated strong detection performance on widely used benchmark datasets
that feature relatively static and constrained human speaking settings~\cite{khalid2021fakeavceleb, cai2024av, rossler2019faceforensics++, cai2023glitch},
often reporting ROC-AUC scores above 0.95.
However, their robustness under unconstrained real-world conditions remains less explored.

\paragraph{Datasets\footnote{DF24 is composed of 300 training and 180 test videos; RW consists of 330 test videos, and \textit{DF-Dynamic} includes 40 curated test videos.}} 
We use Deepfake-Eval-2024 (DF24)~\cite{chandra2025deepfake} and AuViRe RealWorld (RW)~\cite{koutlis2026auvire} to evaluate deepfake detection under unconstrained and dynamic conditions.
Both datasets are collected from media platforms and include both static and dynamic videos, although RW is predominantly static. They also span diverse recording environments,
including varying camera motion, environmental noise, and editing artifacts.
To ensure reproducibility, we exclude samples with missing modalities or invalid links.
All models are trained on the DF24 training set (fine-tuning existing detectors~\cite{liang2024speechforensics,koutlis2026auvire} and training ours from scratch) and evaluated on the DF24 test split (in-domain) and the RW test set (cross-dataset).
In addition, to analyze scenarios where physical cues are most informative, we curate a dynamic subset, \textit{DF-Dynamic}, by manually selecting videos with noticeable speaker motion from both test sets. 

\begin{table}[t]
\small
\centering
\renewcommand{\arraystretch}{0.8}
\caption{\textbf{Results
on the dynamic subset DF-Dynamic
}
This subset contains videos with strong dynamics, such as substantial speaker motion. Under these conditions, our method substantially outperforms existing lip-sync-based detectors.}
\label{tab:df_dynamic}
\vspace{-2mm}
\setlength{\tabcolsep}{6pt}
\resizebox{0.5\linewidth}{!}{\begin{tabular}{lc}
\toprule
\textbf{Method} & \textbf{ROC-AUC} \\
\midrule
SpeechForensics~\cite{liang2024speechforensics} & 0.5096 \\
AuViRe~\cite{koutlis2026auvire} & 0.5517 \\
\textbf{Ours} & \textbf{0.7449} \\
\bottomrule
\vspace{-8mm}
\end{tabular}}
\end{table}

\begin{table}[t]
\small
\centering
\renewcommand{\arraystretch}{0.7}
\caption{\textbf{Results
on the in-the-wild benchmarks.
}
Models are trained on the DF24 training split and evaluated under two settings:
(i) \textit{in-domain} evaluation on the DF24 test set (mixed; static and dynamic videos) and
(ii) \textit{cross-dataset} evaluation on the RW dataset (largely static videos).
(C) reports score-averaging ensembles between lip-sync-based detectors and ours.}
\vspace{-2mm}
\label{tab:in_the_wild}
\resizebox{1.0\linewidth}{!}{
\begin{NiceTabular}{lll}
\toprule
\textbf{Setting} & \textbf{In-domain} & \textbf{Cross-dataset} \\
& \footnotesize{Train: DF24} &
\footnotesize{Train: DF24} \\
 & \footnotesize{Test: DF24 (mixed)} & \footnotesize{Test: RW (static-heavy)} \\
\midrule
\rowcolor{baselineColor}\Block[l]{1-3}{\textbf{(A) Lip-sync-based detectors}}\\
SpeechForensics~\cite{liang2024speechforensics} & 0.6957 & 0.7819 \\
AuViRe~\cite{koutlis2026auvire} & 0.6890 & 0.6430 \\

\midrule
\rowcolor{oursColor}\Block[l]{1-3}{\textbf{(B) Physics-aware detector}}\\
Ours & 0.7329 & 0.6510 \\

\midrule
\midrule

\rowcolor{ensColor}\Block[l]{1-3}{\textbf{(C) Ensemble with ours}}\\
Ours \& SpeechForensics~\cite{liang2024speechforensics} & \textbf{0.7786} & \textbf{0.7971} \\
Ours \& AuViRe~\cite{koutlis2026auvire} & 0.7797 & 0.6634 \\
\bottomrule
\end{NiceTabular}
}
\vspace{-5mm}
\end{table}

\begin{figure}[t] 
  \centering
  \includegraphics[width=1\linewidth]{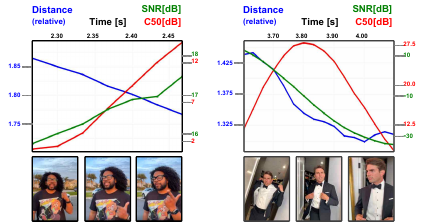}
    \vspace{-5mm}
    \caption{\textbf{Qualitative examples of authentic (left) and deepfake (right) videos.}
    We visualize the temporal trends of \textcolor{distcolor}{distance},
    \textcolor{snrcolor}{SNR}, and \textcolor{c50color}{$C_{50}$}. 
    In authentic videos, the acoustic measures increase as the speaker moves closer to the camera, following the expected physical relationship between distance and speech energy.
    In contrast, deepfake videos often exhibit inconsistent trends between distance and acoustic measures.}
    \vspace{-6mm}
  \label{fig:qualitative}

\end{figure}

\vspace{1mm}
\subsection{Results and Analysis}
\paragraph{Quantitative results} We first evaluate on the curated dynamic subset, \textit{DF-Dynamic}, where physical cues are become particularly informative.
Table~\ref{tab:df_dynamic} shows that existing lip-sync-based detectors exhibit limited performance in this setting, while our physics-aware detector achieves a substantially higher ROC-AUC score.
This suggests that when speaker motion and recording conditions vary dynamically, the physical alignment between speaker distance and acoustic behavior provides a more reliable detection signal than lip–speech consistency alone.


We further evaluate our method on full in-the-wild benchmarks~\cite{chandra2025deepfake, koutlis2026auvire} that include both static and dynamic videos. Table~\ref{tab:in_the_wild} reports (i) in-domain results on the DF24 test set and (ii) cross-dataset results on RW. 
On the in-domain DF24 test set, our model achieves a ROC--AUC of $0.7329$, outperforming existing detectors, while achieving comparable performance on the RW dataset, which is predominantly static. 
Because unconstrained real-world conditions often degrade the mouth region, lip–speech alignment cues become less reliable~\cite{liang2024speechforensics,koutlis2026auvire}; in contrast, our physics-aware approach maintains strong performance by leveraging a fundamentally different signal. 
Moreover, combining our model with existing detectors via a score-averaging ensemble consistently improves performance across both datasets (Table~\ref{tab:in_the_wild}-(C)). These results indicate that the proposed physical-consistency cue is favorable as a standalone signal, and that it provides complementary information that makes ensembling with lip-sync-based detectors even more effective.

\paragraph{Qualitative results}
We visualize authentic and deepfake examples to show how acoustic measures vary with 
estimated speaker distance (see Fig.~\ref{fig:qualitative}). 
In the authentic video, the acoustic measures follow the expected inverse trend: as the speaker moves closer to the camera, both SNR and $C_{50}$ increase due to stronger speech energy and clearer direct sound.
In contrast, deepfake videos often violate this physical relationship, exhibiting inconsistent distance-acoustic trends.
These examples illustrate that the proposed acoustic features provide physically meaningful signals for distinguishing authentic and deepfake videos.

\subsection{Ablation Studies}
\paragraph{Impact of physical features}
Table~\ref{tab:ablation_detailed} reports the detection performance when subsets of physical features are removed.
Removing the SNR-based statistics ($Var_{\textrm{SNR}}, \mathrm{H}_{\textrm{SNR}}$) decreases performance on DF24 ($-0.0262$ AUC), while removing the $C_{50}$-based statistics ($Var_{C_{50}}, \mathrm{H}_{C_{50}}$) causes the largest degradation, reducing the ROC-AUC on DF24 ($-0.1169$) and RW ($-0.042$).
We further analyze the contribution of the statistical descriptors. Removing variance-related features ($Var_{\textrm{SNR}}, Var_{C_{50}}$) leads to a noticeable drop, while removing entropy-based features ($\mathrm{H}_{\textrm{SNR}}, \mathrm{H}_{C_{50}}$) also degrades performance across datasets.
Removing the variance ratio feature $r_{var}$ results in an additional performance decrease ($-0.0193$ AUC on DF24), indicating that the relative variability between acoustic measures provides complementary information.
Overall, best performance is achieved when all physical features are jointly utilized.




\begin{table}[t]
\centering
\caption{\textbf{Ablation of physical features.}
Each row removes a subset of features from our model; $\Delta$ is relative score to ours.}
\vspace{-3mm}
\label{tab:ablation_detailed}
\small
\renewcommand{\arraystretch}{0.9}
\resizebox{\columnwidth}{!}{
\begin{tabular}{@{}lcccc@{}}
\toprule
\multirow{2}{*}{\textbf{Configuration}} &
\multicolumn{2}{c}{\footnotesize \textbf{DF24}} &
\multicolumn{2}{c}{\footnotesize \textbf{RW}} \\
\cmidrule(lr){2-3}\cmidrule(lr){4-5}
& \textbf{ROC-AUC} & \textbf{$\Delta$} & \textbf{ROC-AUC} & \textbf{$\Delta$} \\
\midrule
\textbf{Ours (Full)} & \textbf{0.7329} & -- & \textbf{0.6510} & -- \\
\midrule

\multicolumn{5}{l}{\textbf{(A) Acoustic-measure ablation}} \\
\quad\quad w/o SNR stats ($Var_{\textrm{SNR}}, \mathrm{H}_{\textrm{SNR}}$) & 0.7187 & -0.0262 & 0.6528 & +0.0018 \\
\quad\quad w/o $C_{50}$ stats ($Var_{C_{50}}, \mathrm{H}_{C_{50}}$) & 0.6160 & -0.1169 & 0.6090 & -0.0420 \\
\midrule

\multicolumn{5}{l}{\textbf{(B) Statistical-descriptor ablation}} \\
\quad\quad w/o variance stats ($Var_{\textrm{SNR}}, Var_{C_{50}}$) & 0.6646 & -0.0683 & 0.6246 & -0.0264 \\
\quad\quad w/o entropy stats ($\mathrm{H}_{\textrm{SNR}}, \mathrm{H}_{C_{50}}$) & 0.7097 & -0.0232 & 0.6021 & -0.0489 \\
\midrule

\multicolumn{5}{l}{\textbf{(C) Cross-stat feature}} \\
\quad\quad w/o $r_{var}$ & 0.7136 & -0.0193 & 0.6495 & -0.0015 \\
\bottomrule
\end{tabular}
}
\vspace{-2mm}
\end{table}

\paragraph{Effectiveness of universal sound separation}
We evaluate the impact of the sound separation model~\cite{shi2025sam} by comparing detection performance with and without SAM-audio (see Table~\ref{tab:uss_ablation}).
Using raw audio yields ROC-AUC scores of $0.5945$ on DF24 and $0.5671$ on RW, whereas applying SAM-audio improves them to $0.7329$ and $0.6510$, respectively.
This indicates that separating speech from environmental noise enables more reliable acoustic statistics for detection.

\begin{table}[t]
\footnotesize
\centering
\renewcommand{\arraystretch}{0.4}
\caption{\textbf{Ablation of the sound separation module.}
Removing the sound separation module degrades ROC-AUC scores.}
\vspace{-3mm}
\label{tab:uss_ablation}
\begin{NiceTabular}{lcc}
\toprule
\textbf{Configuration} & \textbf{DF24} & \textbf{RW} \\
\midrule
\textbf{Ours (with SAM Audio)} & \textbf{0.7329} & \textbf{0.6510} \\
\quad w/o SAM Audio & 0.5945 & 0.5671 \\
\bottomrule
\end{NiceTabular}
\vspace{-6mm}
\end{table}

%% file: 4_Conclusion.tex
In this work, we introduce a physics-aware framework for audio-visual deepfake detection in dynamic speaking videos. 
While most existing detectors rely on lip–speech synchronization, which becomes unreliable in dynamic in-the-wild scenarios, we instead leverage a fundamental acoustic constraint linking speaker distance and speech energy.
Our method estimates speaker distance from video and extracts distance-dependent acoustic measures from audio to detect inconsistencies in this relationship. Experiments on curated dynamic subsets show that our approach is particularly effective for dynamic speaking videos. 
Moreover, our method complements lip-sync-based detectors through a simple ensemble, consistently improving detection performance across datasets.




\paragraph{Limitations and future work}
Microphones placed close to the speaker can weaken the distance–speech relationship, and rapidly changing background noise can destabilize physics-aware acoustic measurements. Errors in depth and face estimation can degrade final performance. Our framework is speech-agnostic and may extend beyond human speech to general object-level scenarios that follow similar acoustic propagation principles.





%% file: 5_Acknowledgements.tex
This work was supported by IITP grant (No. 2022-0-00124, No.RS-2022-II220124, Development of Artificial Intelligence Technology for Self-Improving Competency-Aware Learning Capabilities; No. RS-2024-00457882, National AI Research Lab Project; No.RS-2025-25443318, Physically-grounded Intelligence: A Dual Competency Approach to Embodied AGI through Constructing and Reasoning in the Real World) funded by the Korea government(MSIT), the InnoCORE program of the Ministry of Science and ICT(26-InnoCORE-01), and KAIST Undergraduate Research Program (URP).

%% file: 6_Generative_AI_Disclosure.tex
Parts of the manuscript were reviewed with the assistance of Gemini and ChatGPT; all AI-generated suggestions were meticulously reviewed by the authors, who take full responsibility for the content of the publication.